# A Resilience Recovery Method for Complex Traffic Network Security Based on Trend Forecasting

**Sheng Hong,[1] Tianyu Yue,[1] Yang You,[2] Zhengnan Lv,[3] Xu Tang,[2] Jing Hu,[3] and Hongwei Yin[1]**

[1] *School of Cyber Science and Technology, Beihang University, Beijing 100191, China*
[2] *NSFOCUS Technologies Group Co., Ltd., Beijing 100089, China*
[3] *China Electronics Technology Taiji Group Corporation Limited, Beijing 100083, China*

Correspondence should be addressed to Sheng Hong; shenghong@buaa.edu.cn

Due to the rapid development of information technology, a huge and complex traffic network has been established across various sectors, including aviation, aerospace, vehicles, ships, electric power, and industry. However, because of the complexity and diversity of its structure, the complex traffic network is vulnerable to being attacked and faces serious security challenges. Therefore, this paper innovatively proposes a traffic network resilience recovery method based on resilience trend forecasting. In this paper, the risk value is introduced into the analysis of the network fault propagation process, and the Susceptible, Infectious, Recovered, Dead-Risk (SIRD-R) fault propagation model is established. The resilience model of traffic network, which encompasses real-time resilience and overall resilience, is constructed through the integration of network resilience bearing capacity and resilience recovery capacity. Ten, the resilience of complex traffic networks is forecasted by using long short-term memory networks, and the resilience recovery strategy of complex traffic networks based on forecasting is proposed. Finally, the effectiveness and scalability of the proposed method are demonstrated through experimental analysis conducted on a diverse range of complex traffic networks, affirming its applicability in real-world scenarios.

**Keywords:** complex traffic network; network attack; network resilience; network resilience forecasting; recovery strategy

## 1. Introduction

In recent years, the rapid advancement of computer and communication technology has led to a widespread pursuit of intelligence and digitalization as key developmental objectives across various disciplines. In aviation, aerospace, ships, vehicles [1], electric power, industrial production, and other fields, information and network construction have been carried out. Based on the complex and changeable environment scenarios faced by different fields, the complex traffic network is constructed for research and analysis. As a representation of national critical infrastructure, the constructed complex traffic network has emerged as the foremost priority in the construction of national infrastructure and is considered vital to both national security and development [2]. Consequently, it has become a prime target for state-level hacker attacks [3]. In December 2015, hackers planted a virus in the Ukrainian network, paralyzing the country's national traffic network. In May 2017, more than 150 countries, including Ukraine, Russia, Spain, France, and the United Kingdom, suffered WannaCry ransomware attacks, damaging traffic networks and other important industrial facilities. Between March 2019 and May 2020, Venezuela was hit by multiple cyberattacks that paralyzed large areas of its national traffic network and brought the country to near collapse. After the traffic network is attacked, its failure state continues to spread in the network, resulting in the continuous transmission and infection of neighboring nodes, and eventually the overall collapse of the network, which has a great impact on the national social economy.

In recent years, due to the lack of security prevention of the complex traffic network, there are numerous cases of network node equipment failure caused by external attacks, resulting in network collapse and serious harm, facing great security challenges. The security risks faced by the traffic network have seriously threatened human production and life, restricted the development of the social economy, and become one of the major issues related to the national economy and people's livelihood.

The traffic network is a large complex network, and its equipment systems show complex cross-interconnection characteristics. In the event of virus attacks causing the failure of one or more device nodes within the network, this state of failure can propagate to other network nodes through their interconnections, ultimately leading to a complete breakdown of the entire network. Therefore, it is imperative to investigate the propagation of failures in traffic networks following virus attacks and develop optimization methods for enhancing network resilience. This will serve to bolster security against virus attacks, mitigate economic losses, and prevent potential major accidents.

The rest of this paper is structured as follows. Section 2 introduces the related works of network resilience research. Section 3 presents the methods of the trafc network resilience modeling, prediction, and optimization. Section 4 analyzes the results of resilience modeling and optimization of trafc networks through practical case experiments. Section 5 summarizes the full text.

## 2. Related Works

Faced with the security challenges of the traffic network, researchers proposed the concept of network resilience, which refers to a network's ability to deliver and sustain an acceptable level of service in the event of an attack [4]. The term "resilience" originates from Latin, signifying rebound, and was initially employed within materials science [5]. Subsequently, its application has been progressively broadened across various industries. In the face of increasingly complex and frequent network attacks, network resilience has been used in the security research of traffic networks [6]. Network resilience not only represents the ability of the network to cope with attacks, but also reflects the ability of the network to respond to the performance degradation caused by attacks and the ability of the network to recover the normal working state after attacks [7].

In recent years, domestic and foreign researchers have made some progress in the study of network resilience measurement modeling. Li et al. used the ratio of system performance within the maximum allowable recovery time of network interruption to normal state performance as a measure of network resilience [5]. Galinec et al. incorporated the value of system security performance into the model and built a network resilience measurement model by evaluating the known and unknown of related parameters [8]. Based on the barrel theory, Ahmadian et al. took the resilience of the lowest component as the resilience of the network and built a network resilience measurement model [9]. Wang et al. used topological information such as network node degree distribution, link degree distribution, and connection weights to establish a quantitative model of network resilience [10]. Zhang et al. used data such as network load data and load change rate to give a quantitative expression of network resilience and verify it [11]. Ceequeti et al. studied the propagation of attacks through the connection mode among network nodes and proposed a measurement method of network resilience [12]. Wang et al. conducted research from the perspective of physical laws and combined three dimensions of robustness, adaptability, and recoverability in network resilience modeling to build a traffic network resilience measurement model [13].

In addition, many researchers have paid much attention to the recovery of network resilience, which is mainly divided into two aspects: prevention and restoration [14]. In terms of prevention, Imteaj et al. proposed an index of network demand satisfaction rate, which was combined with the index to redistribute network node load, thereby enhancing network resilience [15]. Wang et al. reconstructed network nodes based on backward/forward traffic calculation to improve network resilience [16]. Liu et al. built a network weak dependence model to improve network resilience [17]. Bartos et al. estimate the attack probability and build a blacklist for reconnaissance and monitoring, so as to prevent the deterioration of network resilience [18]. In terms of recovery, Amini et al. realized the resilience recovery of the single-lattice network by rebuilding the failed connection of nodes [19]. Hong et al. proposed a method for random maintenance of all failed nodes according to a certain probability of repair [20]. Gong et al. combined the node load recovery coefficient to form a recovery scheme considering the resilience distance and realized the optimal recovery of network resilience [21]. Laishram et two attributes, core strength and core influence, and proposed the kCore maximum resilience algorithm to improve network resilience [22].

However, there are still some problems that need to be solved in the study of network resilience of traffic networks. First of all, the existing network resilience measurement methods are only an incomplete and inadequate evaluation of network resilience. Secondly, the current recovery strategies on the network resilience of the traffic network are usually analyzed from a certain aspect, lacking comprehensive research and analysis.

In order to address the aforementioned issues, this paper conducts resilience modeling, forecasting, and recovery research on the traffic network. It establishes the traffic network SIRD-R failure propagation model and network resilience model, forecasts the resilience change trend of the traffic network, and proposes the resilience recovery strategy of the traffic network based on forecasting.

## 3. Resilience Modeling and Optimization Method of Traffic Networks

*3.1.* Traffic *Network Failure Propagation Analysis.* When Certain nodes within the traffic network are subjected to attacks and subsequently fail. The resulting impact on network connectivity can trigger a chain reaction, leading to the propagation of local failures throughout the entire network. This cascade effect may ultimately result in widespread functional impairment or even complete collapse of the network, a phenomenon commonly referred to as cascade failure propagation [23]. The cascade failure propagation process of the network is shown in Figure 1.

Figure 1(a) shows the initial state of the network. When the network is attacked, one of the nodes is in a failed state, as shown in Figure 1(b). Due to the failure of this node, the function of the node cannot be carried out normally, and the adjacent node is affected by this node and also fails, as shown in Figure 1(c). Similarly, the failure state of a node further spreads to neighboring nodes, and finally, the overall function of the entire network fails and the network crashes, as shown in Figure 1(d).

The SIRD model is employed to examine the process of failure propagation within the traffic network, with the node devices categorized into four distinct states [24]. Failure state I: The node in this state has failed and has the ability to propagate the failure state to other nodes. Susceptible state S: At present, the node device in this state has not been propagated failure, but there is a certain possibility of failure propagation. Recovery state R: The node device in this state recovers from the failure state due to the security protection and recovery means of the device itself. Breakdown state D: In this state, the node device cannot be recovered from the failure state, resulting in the collapse of the node device.

The four state transitions are shown in Figure 2.

In Figure 2, the propagation rate of S state to I state is $\beta$, the recovery rate from I state to R state is $\mu$, and the node device collapse rate from I state to D state is recorded as $\lambda$.

The formula of the SIRD model of failure propagation is as follows: at the time interval $\Delta t$, the changes of S, I, R, and $D$ states $\Delta S$, $\Delta I$, $\Delta R$, and $\Delta D$ are as follows:

$$\begin{aligned} \frac{\Delta S}{\Delta t} &= -\beta S, \\ \frac{\Delta I}{\Delta t} &= \beta S - \mu I - \lambda I, \\ \frac{\Delta R}{\Delta t} &= \mu I, \\ \frac{\Delta D}{\Delta t} &= \lambda I. \end{aligned} \quad (1)$$

When $\Delta t \longrightarrow 0$, the above equation can be rewritten as a differential equation.

$$\begin{aligned} \frac{dS}{dt} &= -\beta S, \\ \frac{dI}{dt} &= \beta S - \mu I - \lambda I, \\ \frac{dR}{dt} &= \mu I, \\ \frac{dD}{dt} &= \lambda I. \end{aligned} \quad (2)$$

In the SIRD failure propagation model of traffic networks, the state conversion rate of each node in the network is assumed to be consistent, with uniform values for $\beta$, $\mu$, and $\lambda$. However, in real-world traffic networks, variations in node characteristics such as location, type, and security measures result in differing propagation rates ($\beta$), recovery rates ($\mu$), and crash rates ($\lambda$) following an attack.

Therefore, node risk value $f$ is introduced in the calculation of the conversion rate of each node state, as shown in Figure 3, and the risk value $f$ is used to represent the impact of the heterogeneity of network nodes on the failure propagation, forming the SIRD-R model, as shown in Figure 4.

In the SIRD-R model, nodes with lower risk values exhibit improved security protection measures, resulting in a decreased failure propagation rate, an increased recovery rate, and a reduced collapse rate. The specific formula is as follows:

$$\begin{aligned} \frac{dS}{dt} &= -f\beta_0 S, \\ \frac{dI}{dt} &= f\beta_0 S - (1 - f)\mu_0 I - f\lambda_0 I, \\ \frac{dR}{dt} &= \mu_0 I, \\ \frac{dD}{dt} &= f\lambda_0 I. \end{aligned} \quad (3)$$

In the formula, $\beta_0$, $\mu_0$, and $\lambda_0$ are defined as the basic propagation rate, recovery rate, and collapse rate of the network. Combined with the risk value $f$ of each node, the analysis of each state of the network node is realized.

*3.2. Trafc Network Resilience Measurement Modeling.* After analyzing the failure propagation process of the traffic network, the resilience of the traffic network is further modeled. The concept of network resilience refers to the capacity of a network to return to its normal state, either through internal adjustments or external interventions, following the emergence and propagation of a failure node subsequent to an attack.

In the study on resilience modeling of traffic networks, the classic model is the resilience triangle model proposed by Bruneau [25]. In subsequent studies, Bruneau's resilience triangle model was further refined and network

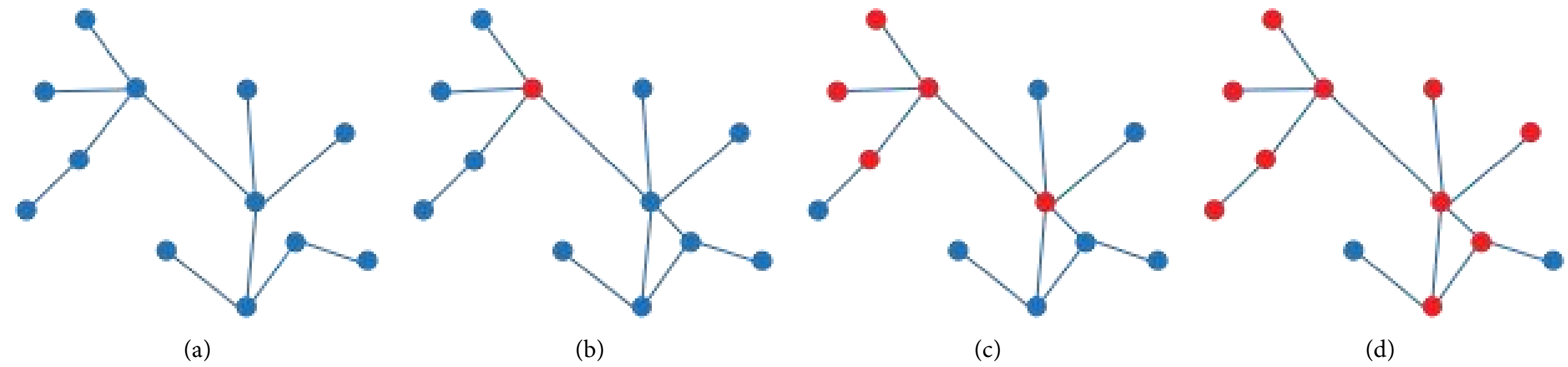

FIGURE 1: Cascade failure propagation process of network. (a) Initial status. (b) Attack causes failed node. (c) Failure propagation. (d) Further spread of failure.

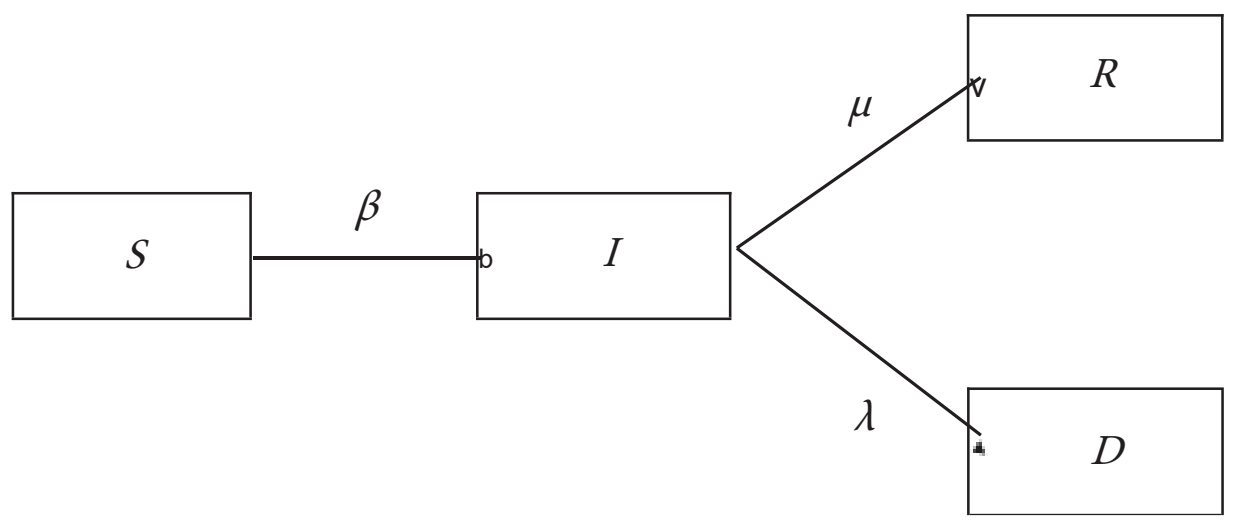

FIGURE 2: SIRD failure propagation model.

Introduced value at risk

S (t) | f | S (t + 1)

I (t) | λ | $f\lambda_0$ | I (t + 1)

$\mu$ | $(1-f)\mu_0$

R (t) | $f\beta_0$ | R (t + 1)

$\beta$

D (t) | D (t + 1)

Node state transition

Failure propagation process

FIGuRE 3: SIRD-R failure propagation model architecture.

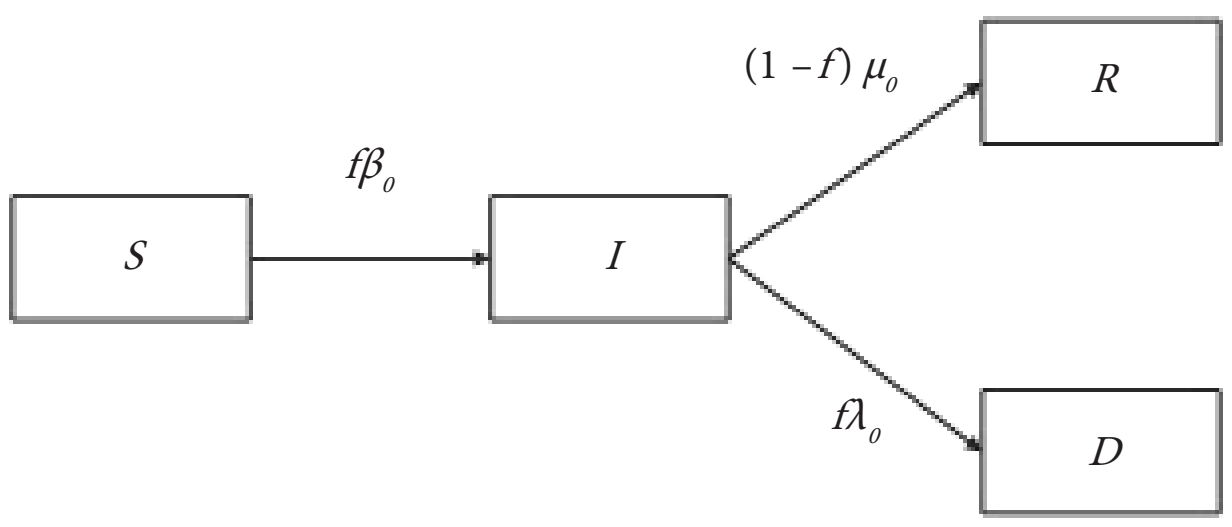

FIGURE 4: SIRD-R failure propagation model.

Resilience was defined as the percentage of total network losses over a period of time [26]. Another network resilience model is the resilience measuring quotient model, which includes fault response and recovery measures [27]. The network resilience triangle model holds that the performance of the network is a cliff-like decline after the attack, but in reality, there is a certain time process when the overall performance of the network drops to the lowest point after the attack. The network resilience quotient model holds that there is a constant minimum level of network performance over a period of time, which does not exist in reality due to constant failure propagation in the network.

Therefore, this paper explores a resilience modeling method, which divides network resilience into two parts: real-time resilience and overall resilience. In the real-time resilience measurement, the resilience of the network at the current moment and the change in resilience are given. In the overall resilience research and analysis, the overall resilience measurement results after the network recovers from the attack are given.

*3.1. Real-Time Resilience.* According to the failure propagation analysis of traffic networks, the state of each node in the network can be obtained. The nodes in the S and R states are not failed or recovered, that is, available nodes. The nodes in the I and *D* states are failed or crashed, that is, unavailable nodes.

Based on previous research, the network performance function $K(t)$ is used to express the working performance status of the traffic network at different times [28]. The calculation formula $K(t)$ is as follows:

$$K(t)=\frac{\sum_{i\in N}W_i(t)\times A_i(t)}{\sum_{i\in N}W_i(t)}, \tag{4}$$

where $N$ is the set of all nodes in the network, $W_i(t)$ is the relative weight of the node at time $t$, and $A_i(t)$ is the available capability of the node under attack failure propagation.

According to the SIRD-R failure propagation model, the formula $A_i(t)$ is as follows:

$$A_i(t)=\begin{cases}0, & I, D\text{ state}\\ 1, & S, R, \text{ state.}\end{cases} \tag{5}$$

The relative weight of a node $W_i(t)$ can be expressed as the degree centrality of the network node.

$$W_i(t) = \frac{1}{N-1} \sum_{j=1}^{N} a_{ij} (i \neq j), \tag{6}$$

where N is the total number of nodes in the network and $a_{ij}$ is the connection edge between nodes ij.

The network performance function K(t) after the traffic network is attacked is obtained through experiments, as shown in Figure 5.

As shown in Figure 6, the ratio of the network performance under attack K(t) to the network performance under normal circumstances $K_0$ (t) at time *t* is taken as the network resilience E (t) at time *t*. The formula is as follows:

$$E(t) = \frac{\int_{T=t}^{t+\Delta t} K(t)}{\int_{T=t}^{t+\Delta t} K_0(t)}. \tag{7}$$

*3.2. Overall Resilience.* The overall resilience of the trafc network is to evaluate the ability of the network to withstand and recover from attacks. As shown in Figure 7, the capability of the network to withstand attacks is shown as the value of real-time resilience at the lowest point, while the recovery capability of the network is shown as the value of real-time resilience when the time approaches infinity.

Therefore, the network resilience bearing capacity of the traffic network $E_{withstand}$ is the minimum value of real -time resilience, and the formula is as follows:

$$E_{withstand} = \min (E(t)). \tag{8}$$

The network resilience recovery capacity $E_{recover}$ is the real-time resilience at time T of the experimental simulation cut-off time, and the formula is as follows:

$$E_{recover} = E(T). \tag{9}$$

The overall resilience of the network *E* can be expressed by the area integral of the resilience curve E (T) during the whole time, calculated by the formula.

$$E = \int_{t=0}^{T} E(t). \tag{10}$$

After *E* is normalized, the above formula becomes as follows:

$$E = \frac{\int_{t=0}^{T} E(t)}{\int_{t=0}^{T} 1}. \tag{11}$$

By further simplifying the above formula, the final formula of the overall resilience *E* of the traffic network is as follows:

$$E = \frac{1}{T} \int_{t=0}^{T} E(t). \tag{12}$$

*3.3.* Traffic *Network Resilience Forecasting.* The evolution of Network resilience varies under different attack scenarios in the traffic network. Therefore, it is necessary to forecast and

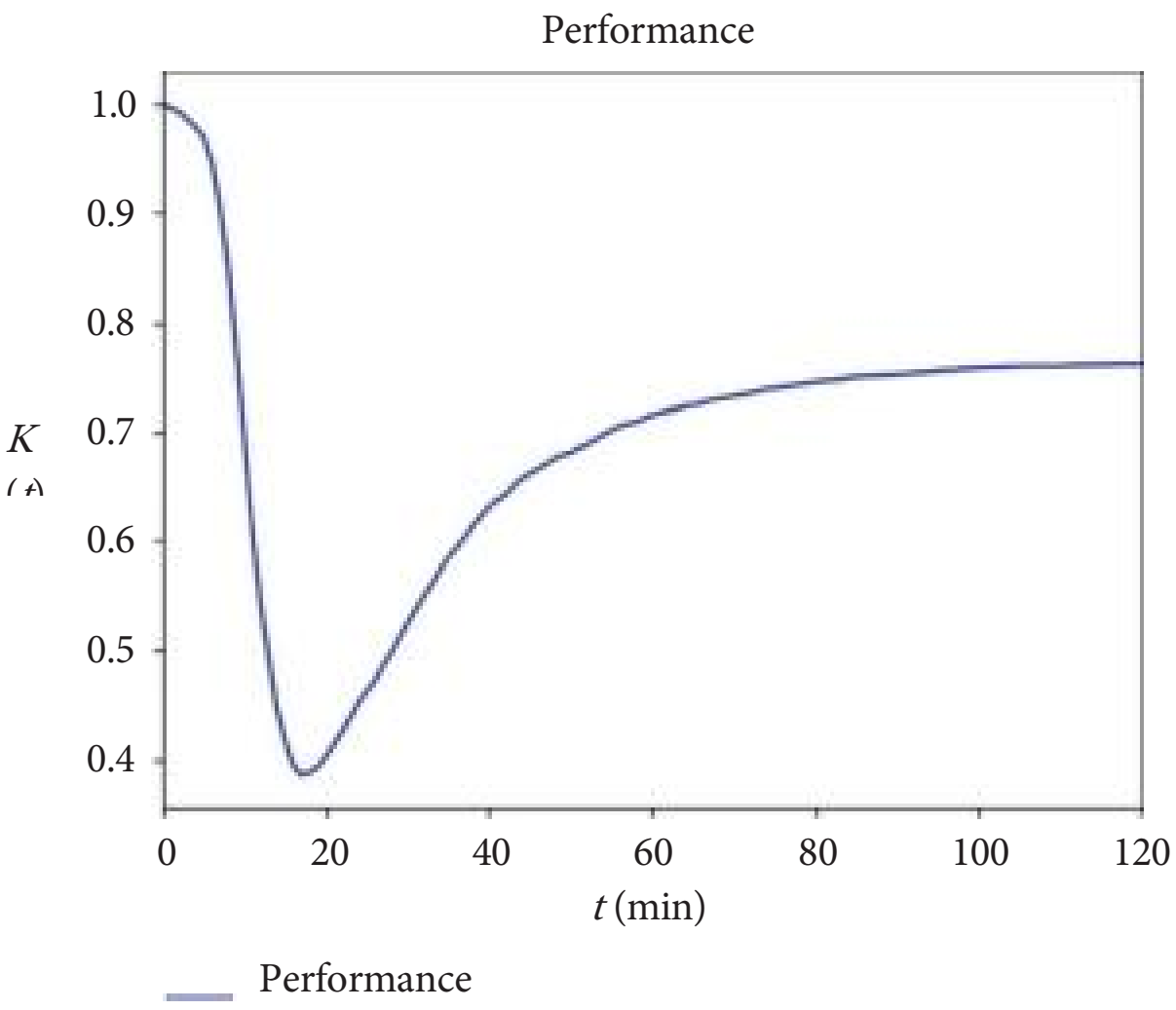

FIGURE 5: Traffic network performance function.

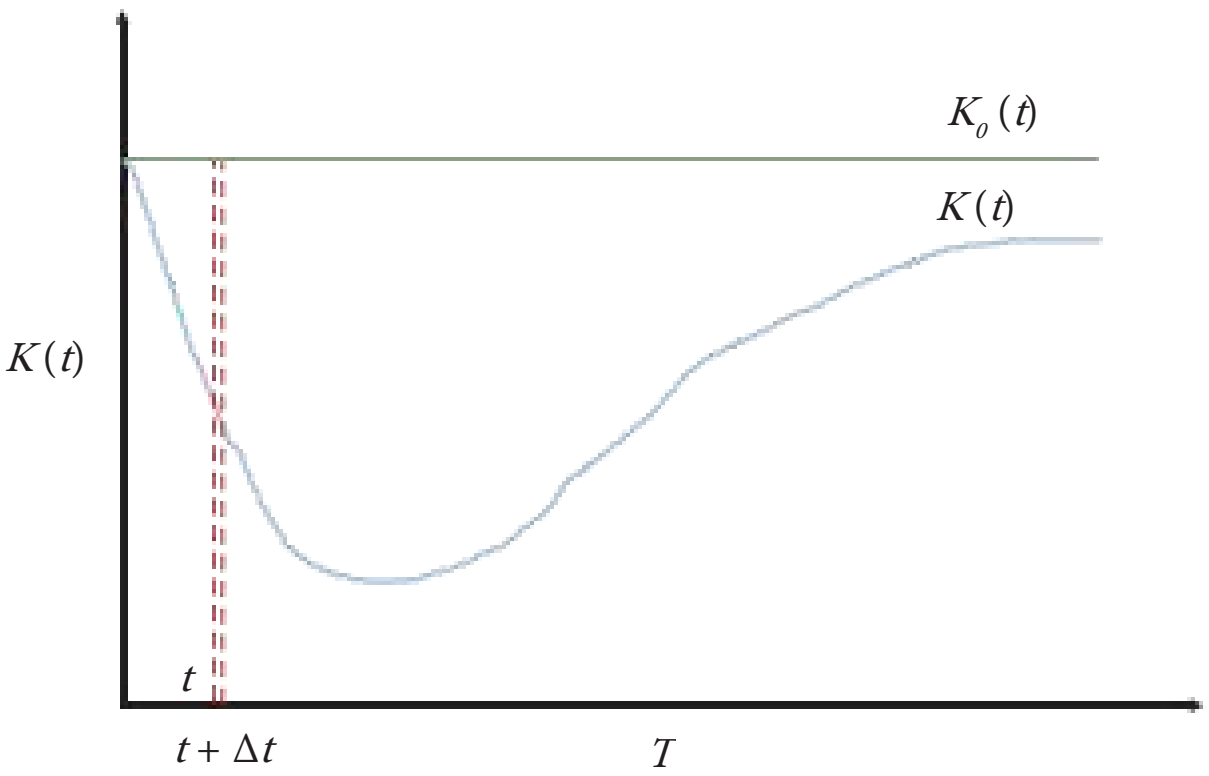

FIGURE 6: Traffic network real-time resilience calculation example.

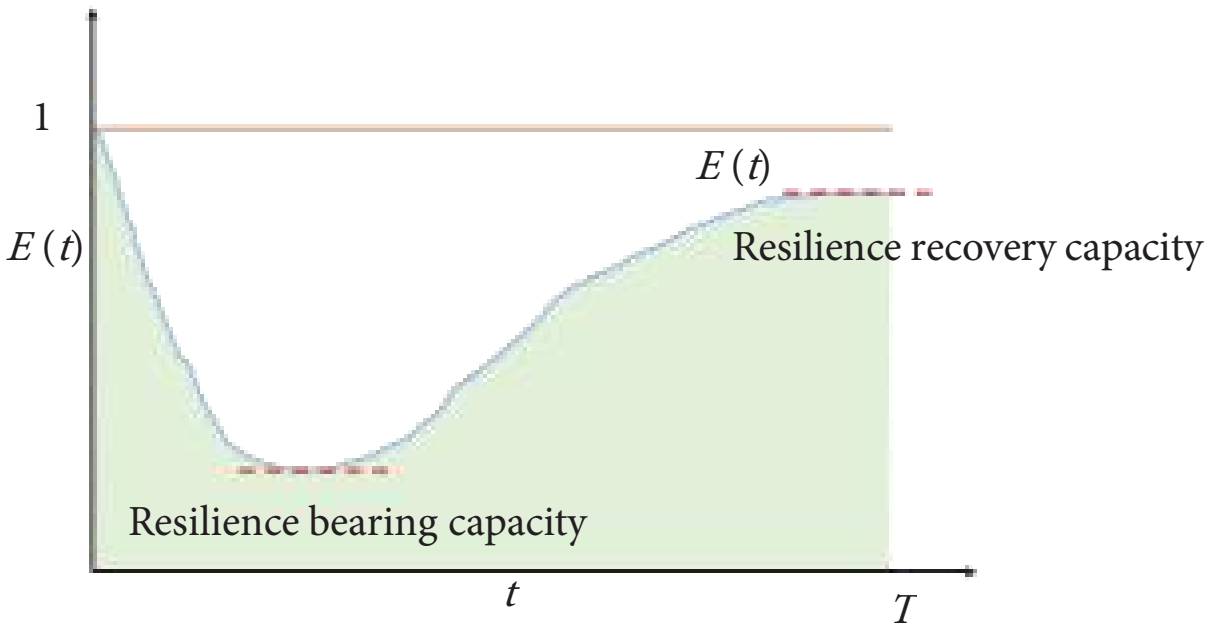

FIGURE 7: Example of overall resilience calculation for traffic network.

Analyze the trend of the traffic network resilience in order to promptly discern any shifts in network resilience and implement timely remedial measures.

The resilience of traffic networks is characterized by dynamic and time-varying data. For such data forecasting with time structure, recurrent neural network (RNN) shows better performance than traditional neural networks because of its internal loop structure, which can retain the previous information and use it in the subsequent calculation. However, due to its long-term deep connections, RNNs will encounter the problem of gradient disappearance during backpropagation training, making RNNs unable to deal with long-term dependent data. Long short-term memory network (LSTM) is an improved RNN model, which introduces the concept of self-loop to generate a path of long-term continuous gradient flow, and introduces a gating mechanism to control the information flow to solve the problem that RNN cannot handle long-term dependence [29].

Therefore, according to the traffic network resilience model, the LSTM algorithm is used to forecast the resilience change trend of the traffic network. Firstly, the resilience data of the traffic network are loaded, the data are preprocessed, the number of observations is set to forecast the next moment each time, and the data are divided into training data and test data. Ten, the LSTM network model parameters are set up, and the LSTM network forecasting model is constructed. Ten, the LSTM network forecasting model is trained. Finally, the target data is forecasted after the LSTM network forecasting model is trained.

*3.4. Resilience Recovery Strategy of* Traffic *Network Based on*

*Forecasting.* During the initial stage of failure propagation in a traffic network attack, the limited number of failure nodes may initially escape notice. However, as failures spread across a wider area, appropriate measures will be implemented and increased attention will be directed toward the situation. Through analysis and forecasting of the dynamic trends in traffic network resilience, it is possible to identify the evolving patterns of network resilience. This enables proactive measures to be implemented in advance for optimizing network resilience and enhancing the overall resilience of the traffic network. The network resilience recovery strategy based on forecasting can be divided into two methods: node protection based on forecasting and node recovery based on forecasting. The traffic network resilience recovery model architecture based on forecasting is shown in Figure 8.

*3.4.1. Node Protection Based on Forecasting.* After fore-By casting the traffic network resilience, protective measures are taken for nodes in the network that have not failed, so that the node risk value is reduced, the failure propagation rate is reduced, the recovery rate is increased, and the collapse rate is reduced, so as to improve the network resilience.

*3.4.2. Node Recovery Based on Forecasting.* Based on the Forecasting of the traffic network resilience, recovery measures are taken to restore the nodes in the failed state to the normal working state, so as to restrain the failure propagation of the nodes in the network and improve the network resilience.

## 4. Experimental Results and Analysis

The traffic network data studied in this paper are Gold Coast, a city in southern Queensland, Australia. The traffic network data contain 4,807 nodes and 11,140 connected edges.

*4.1. Modeling Results and Analysis of Network Resilience.* The SIRD model is used to plot the dynamic changes of four types of nodes under node fault propagation attack in a traffic network. β is defined as 0.3, μ is 0.1, λ is 0.02, and the failure number of attacked nodes is set to 1 at the initial moment. The results are shown in Figure 9.

As can be seen from Figure 9, the failed node devices in the network first rose rapidly, then slowly fell, and finally, there are no failed nodes in the stable state.

By using the SIRD-R failure propagation model, the dynamic changes of four types of traffic network nodes are drawn. $\beta 0$ is defined as 0.3, $\mu 0$ as 0.1, and $\lambda 0$ as 0.02. The risk value $f$ of each node is generated by random simulation, and the number of failures of attacked nodes at the initial moment is set to 1, as shown in Figure 10.

By comparing Figures 9 and 10, it can be seen that under the same $\beta$, $\mu$, and $\lambda$, the entire failure propagation time process becomes longer due to the introduction of node risk value. Because some nodes have better security protection measures and lower risk value, the peak value of nodes in failure state is reduced. However, some nodes have weaker security protection measures and higher risk value, and the final number of recovered nodes is also reduced.

The real-time resilience of the traffic network is obtained by the method of Section 2, as shown in Figure 11.

As can be seen from Figure 11, in the initial stage of the network attack, the failure state spreads rapidly among nodes, and the real-time resilience of the network decreases rapidly. After a prolonged period of cascading failure propagation, the majority of nodes have transitioned to a failed state, while certain nodes have been able to recover from this state as a result of implemented security protection measures. At this time, the real-time resilience of the network begins to recover. However, because some nodes in the network fail to recover but collapse, the real-time resilience of the network can not return to the resilience performance before the final stability.

Through experiments, the network resilience bearing capacity Ewithstand, network resilience recovery capacity $E_{recover}$, and overall network resilience $E$ of the traffic network are shown in Table 1.

Select different values for the experimental simulation cutoff time $T$, and obtain the results of network resilience bearing capacity $E_{withstand}$, network resilience recovery capability Erecover, and overall network resilience $E$, as shown in Table 2.

It can be seen from Table 2 that with the increase of experimental simulation time, the network resilience recovery ability and the overall network resilience have improved. When the experimental simulation time is relatively short, the improvement of network resilience recovery ability and overall network resilience is more significant, while when the experimental simulation time is longer, the improvement of network resilience recovery ability and overall network resilience continues to decline. Tis is because when the experimental simulation time is short, the entire network failure propagation process of the trafc network cannot be efectively simulated, and the network resilience performance results at this time cannot represent the entire network failure propagation process. With the increase of simulation time, the evolution of the failure propagation process of the network is more perfect, and the resilience performance of the network is closer to the actual situation.

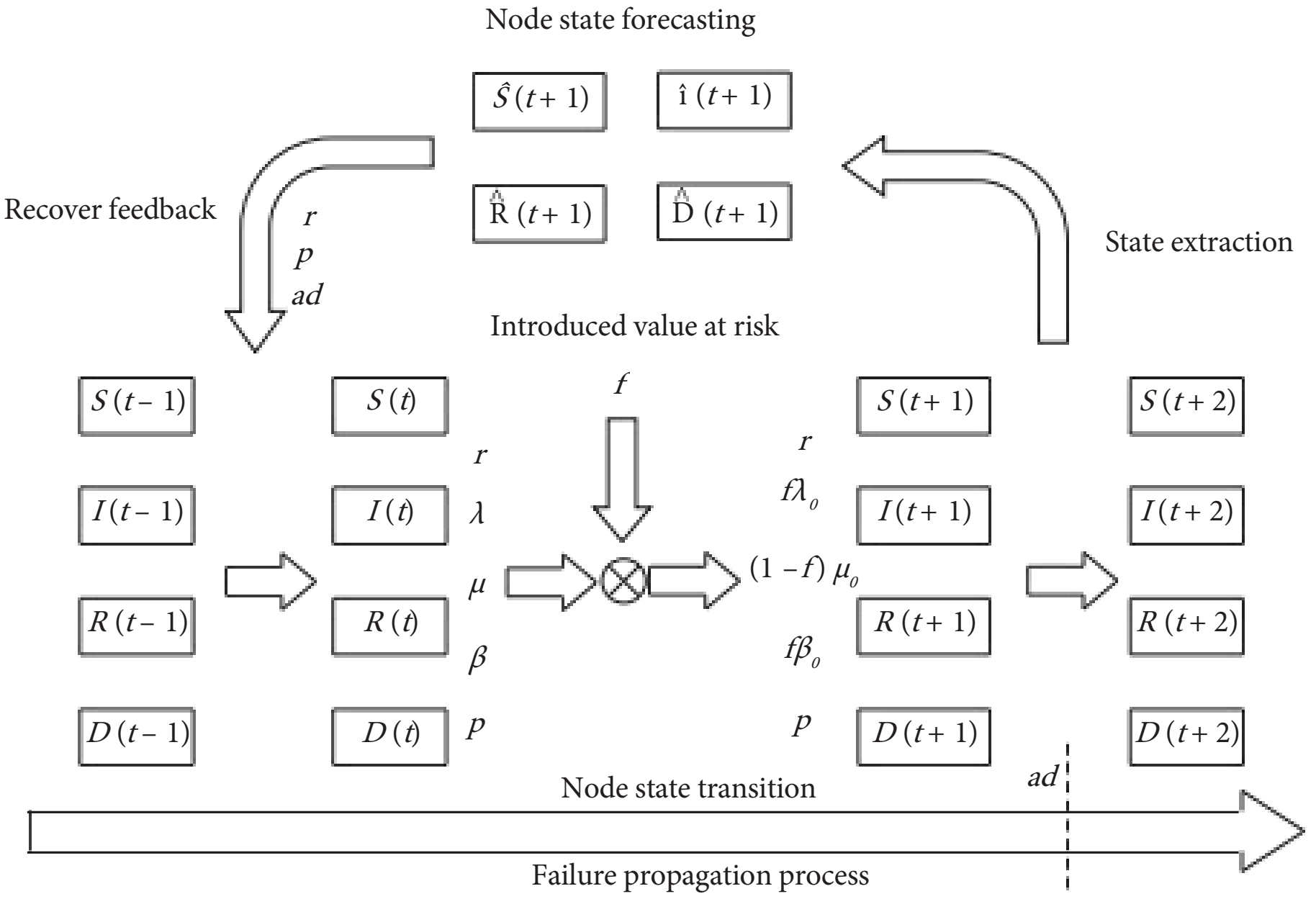

FIG. 8: Resilience recovery model architecture of traffic network based on forecasting.

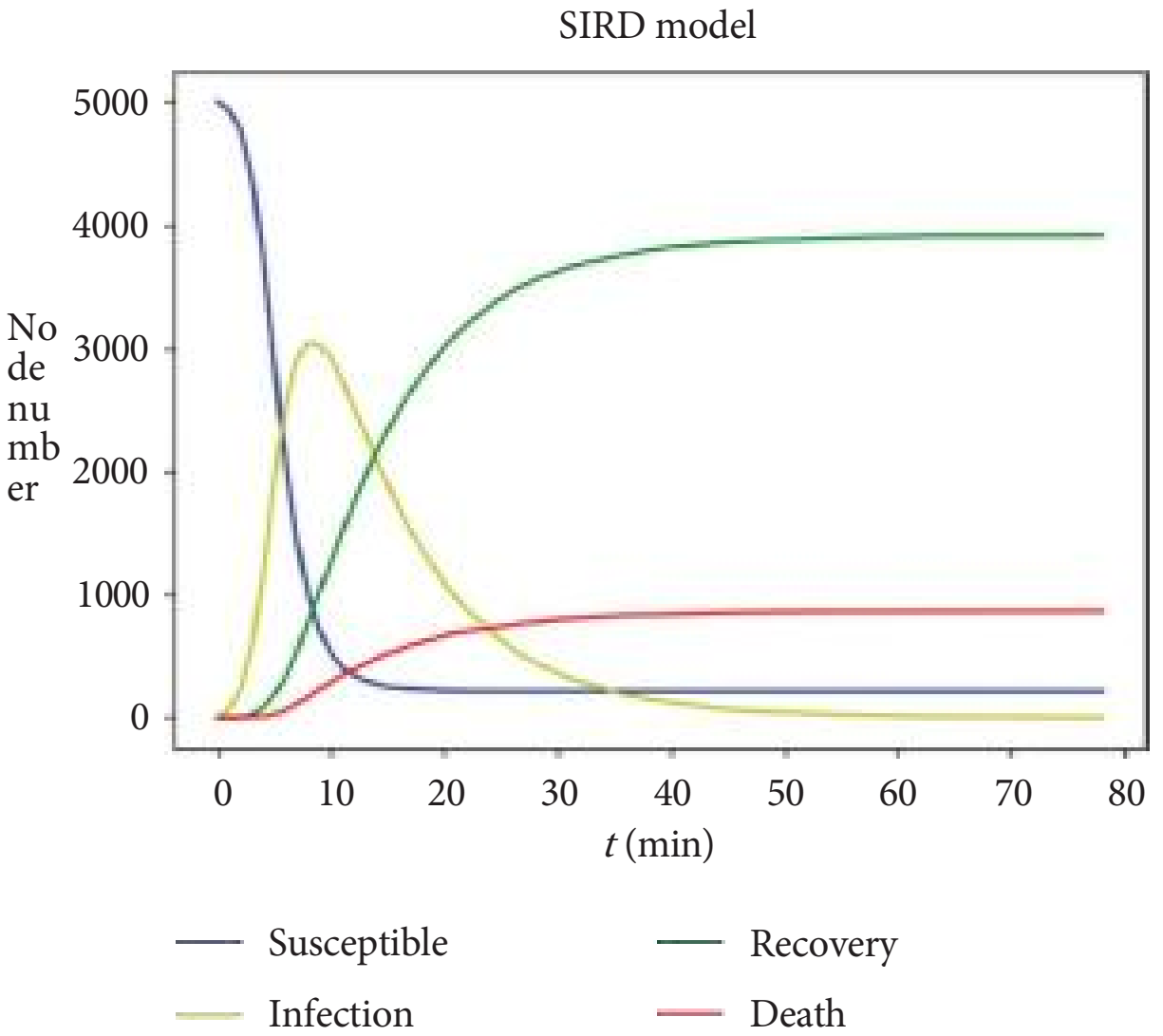

FIGURE 9: SIRD failure propagation model.

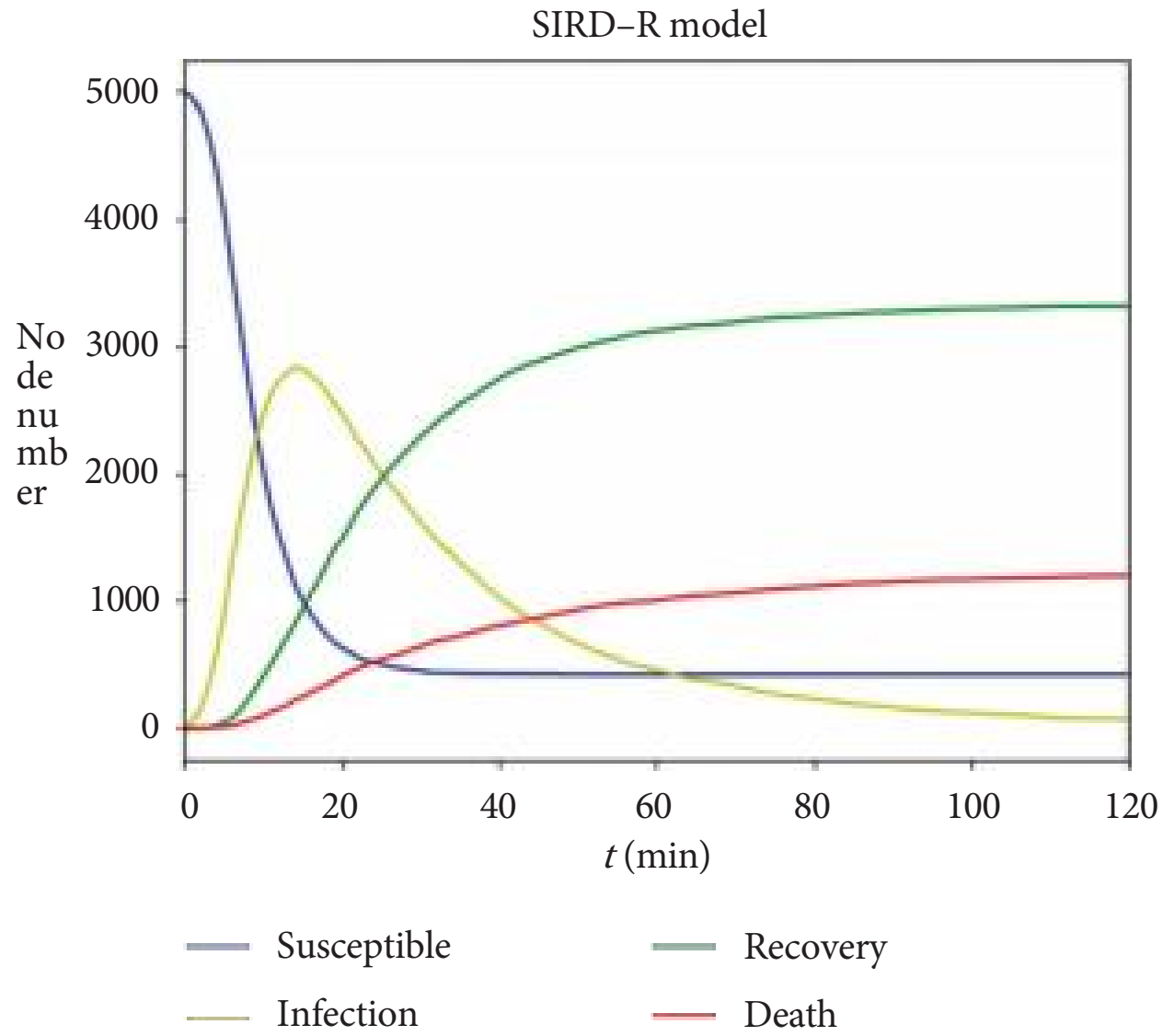

FIGURE 10: SIRD-R failure propagation model.

*4.2. Experimental Analysis of Network Resilience Forecasting and Optimization Results.* LSTM algorithm is used to forecast the traffic network resilience, and the forecasting results are shown in Figure 12.

In order to evaluate the forecasting degree of the LSTM algorithm, the mean relative error (MRE), mean square error (MSE), and R2 determination coefcient are used to evaluate.

MRE is the average of the relative difference between the real value and the forecasted value, which can reflect the error between the forecasted value and the actual value, and its calculation formula is as follows:

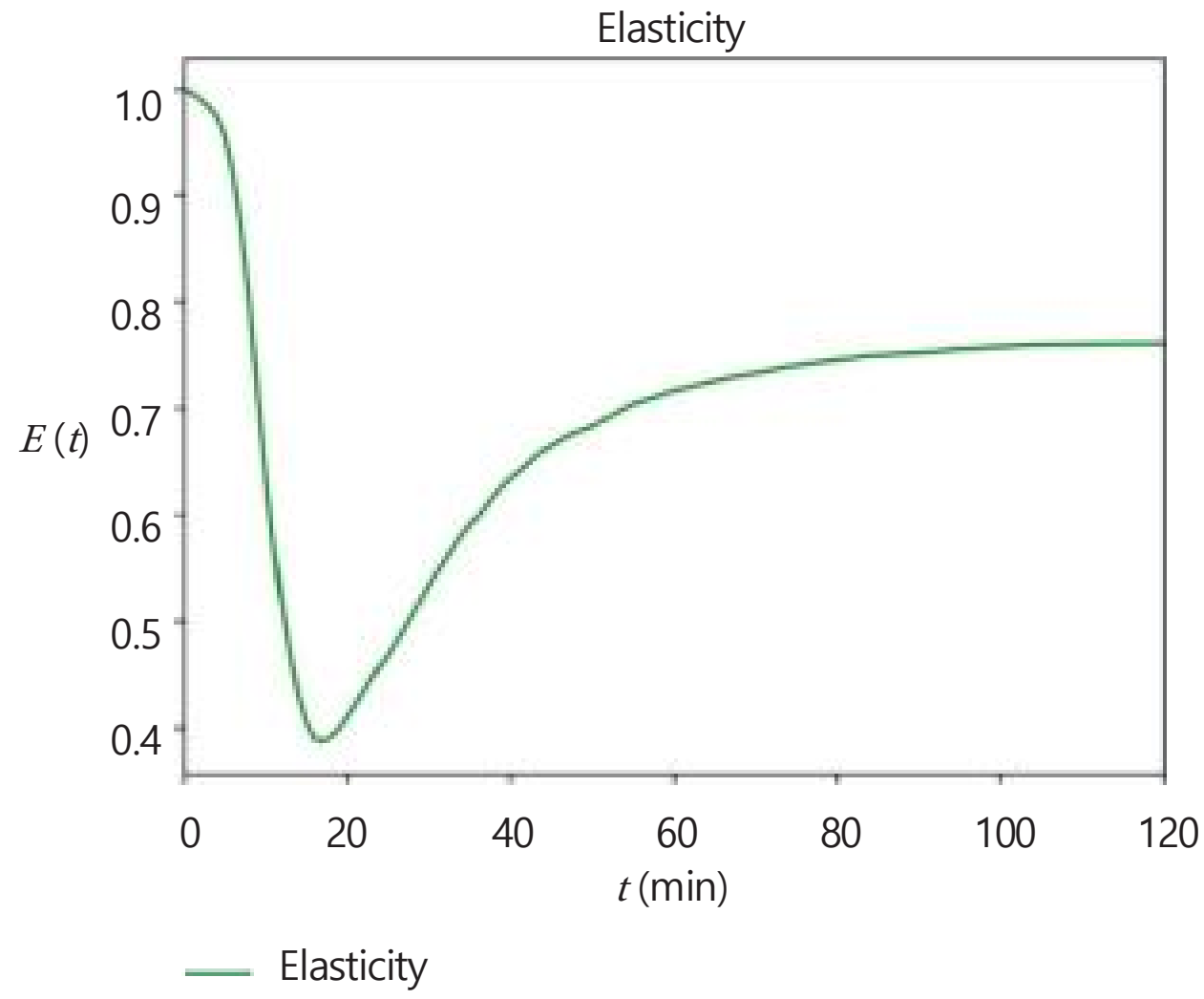

FIGURE 11: Traffic network real-time resilience.

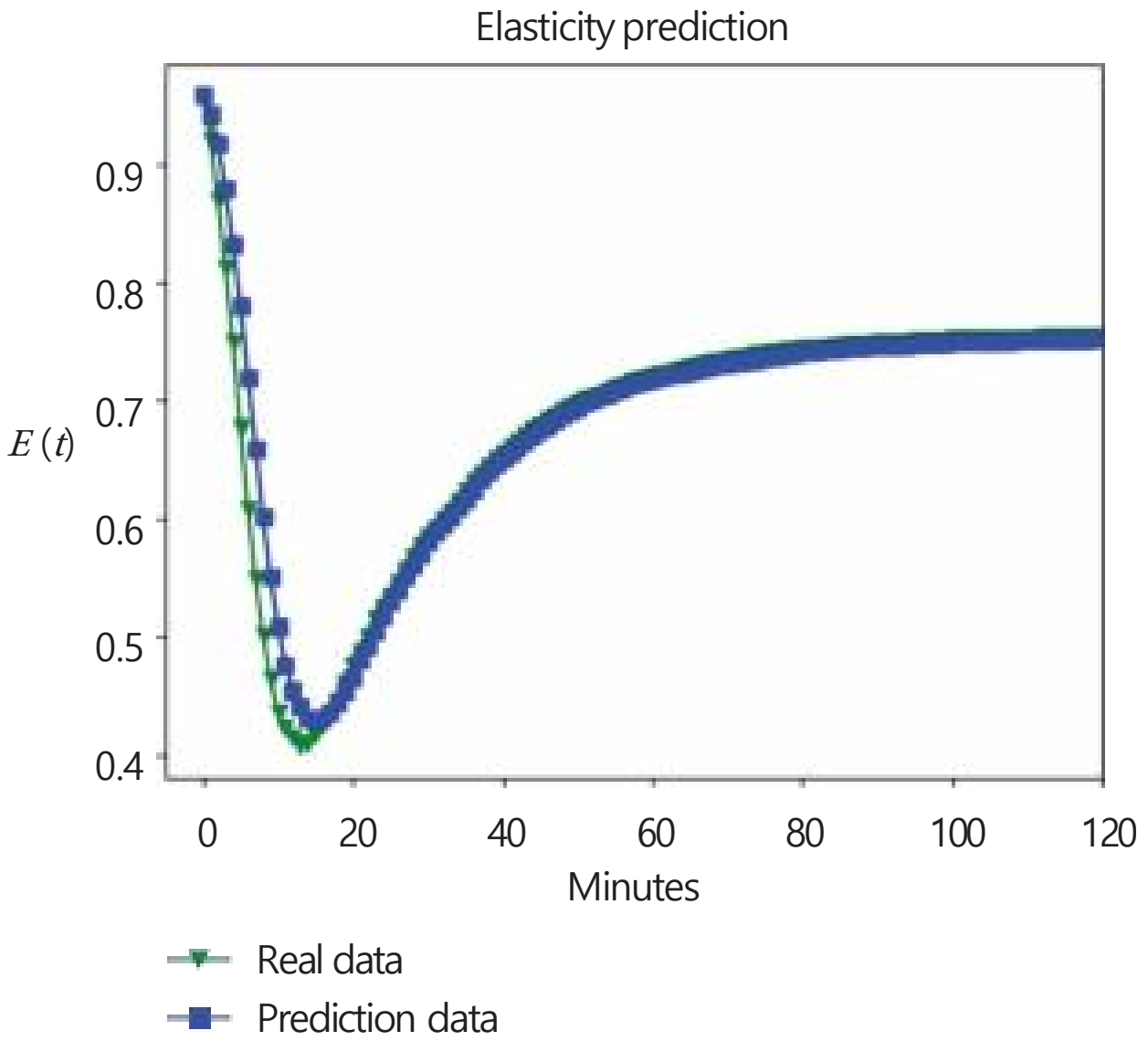

FIGURE 12: Forecasting results of traffic network resilience.

TABLE 1: Traffic network bearing capacity, recovery capacity, and overall resilience.

| $E_{withstand}$ | $E_{recover}$ | $E$ |
|---|---|---|
| 0.4014 | 0.7678 | 0.7474 |

TABLE 2: Overall resilience of trafc network under diﬀerent cutoﬀ time $T$ (the value of $E_{withstand}$ is unchanged at 0.4014).

| $T$ | $E_{recover}$ | $E$ |
|---|---|---|
| 30 | 0.5618 | 0.5725 |
| 50 | 0.6992 | 0.5981 |
| 80 | 0.7517 | 0.6471 |
| 120 | 0.7644 | 0.6847 |
| 160 | 0.7670 | 0.7051 |
| 200 | 0.7676 | 0.7175 |
| 250 | 0.7678 | 0.7276 |
| 300 | 0.7678 | 0.7343 |
| 400 | 0.7678 | 0.7427 |

$$\mathrm{MRE} = \frac{1}{n}\sum_{i=1}^{n}\left|\frac{y_{i0} - y_i}{y_i}\right|, \tag{13}$$

where $y_i$ is the actual resilience result, $y_{i0}$ is the forecasted resilience result, and $n$ is the total number of data.

The MSE is computed as the average of the sum of squares between the forecasted and actual values, serving as a measure to quantify the discrepancy between forecasted outcomes and true values. Its calculation formula can be expressed as follows:

$$\mathrm{MSE} = \frac{1}{n}\sum_{i=1}^{n}\left(y_{i0} - y_i\right)^2. \tag{14}$$

R2 coefficient reflects the accuracy of the forecasted fitting data. The R2 coefficient ranges from 0 to 1. The closer the value is to 1, the better the forecasted data are fitted. The calculation formula is as follows:

$$\mathrm{R2} = 1 - \frac{\sum_{i=1}^{n}\left(y_{i0} - y_i\right)^2}{\sum_{i=1}^{n}\left(\overline{y}_i - y_i\right)^2}, \tag{15}$$

where $\overline{y}_i$ is the mean value of actual resilience results.

MRE, MSE, and R2 determination coefficient are used to evaluate the forecasting accuracy of the LSTM algorithm on traffic network resilience. The evaluation results are shown in Table 3.

As can be seen from Table 3, the MRE is 0.00639, the MSE is 0.00016, and the R2 coefficient is 0.96842. The error of the resilience forecasting results of the traffic network using the LSTM algorithm is very small, and the R2 coefficient is close to 1, so the resilience forecasting result is very good.

CNN, RNN, and LSTM algorithms are, respectively, used to forecast the trafc network resilience, and the resilience forecasting results under diﬀerent algorithms are compared as shown in Figure 13. The forecasting accuracy is shown in Table 4.

It can be seen from Figure 13 and Table 4 that the CNN algorithm has the worst forecasting eﬀect on the trafc network resilience and the largest forecasting error. The forecasting result of the RNN algorithm is better than the CNN algorithm but worse than the LSTM algorithm, and the forecasting error is in the middle position. The resilience forecasting result of the LSTM algorithm is closest to the actual elastic result, and its forecasting error is the smallest. Therefore, using the LSTM algorithm to forecast the traffic network resilience can achieve better expected results.

TABLE 3: Resilience forecasting accuracy of traffic network based on LSTM algorithm.

| MRE | MSE | R2 coefficient |
|---|---|---|
| 0.00639 | 0.00016 | 0.96842 |

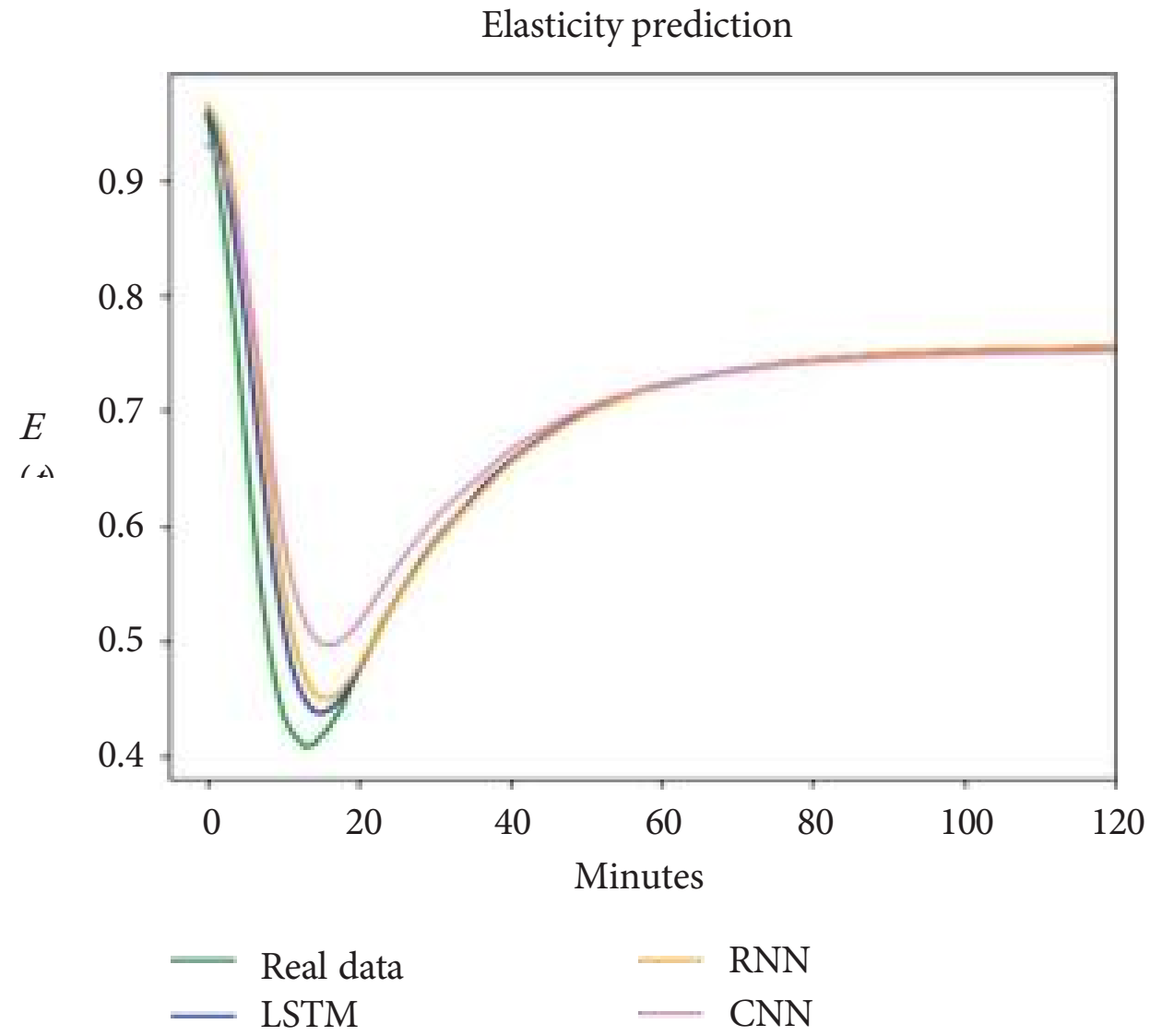

FIGURE 13: Comparison of traffic networks' resilience forecasting results with different algorithms.

TABLE 4: Comparison of traffic networks' resilience forecasting accuracy with different algorithms.

| Algorithm | MRE | MSE | R2 coefficient |
|---|---|---|---|
| LSTM | 0.00639 | 0.00016 | 0.96842 |
| RNN | 0.00898 | 0.00051 | 0.91152 |
| CNN | 0.01859 | 0.00082 | 0.85833 |

The traffic network resilience under different attacks in different scenarios is further forecasted. The forecasting results are shown in Figure 14, and the forecasting accuracy is shown in Table 5.

As can be seen from Figure 14 and Table 5, the resilience forecasting results of the traffic network under different conditions are very close to the actual results, the forecasting errors are very small, and the R2 coefficients are close to 1.

The network resilience of the node protection based on forecasting under different protection rates is shown in Figure 15 and Table 6.

It can be seen from Figure 15 and Table 6 that the node protection strategy based on forecasting can effectively improve the traffic network resilience, and the resilience bearing capacity, resilience recovery capacity, and overall resilience of the traffic network are all improved. And with the improvement of the protection rate, the resilience bearing capacity, resilience recovery capacity, and overall resilience continue to improve. The node protection strategy based on forecasting is mainly to take protective measures for the nodes that have not failed in the network, reduce the possibility of nodes becoming failed, and prevent the spread of network node failures. However, it cannot recover the failed nodes, so the network resilience cannot recover to the initial 100% state. In addition, with the improvement of the protection rate, the time cost is also constantly reduced.

The network resilience of the node recovery based on forecasting under different recovery rates is shown in Figure 16 and Table 7.

It can be seen from Figure 16 and Table 7 that the node recovery strategy based on forecasting can significantly improve the traffic network resilience, and the network resilience bearing capacity, resilience recovery capacity, and overall network resilience have been significantly improved. With the improvement of the recovery rate, the network resilience bearing capacity, resilience recovery capacity, and overall network resilience have been continuously improved. The node recovery strategy based on forecasting is mainly to take recovery measures for failed nodes in the network and restore the nodes from the failed state to the normal working state, so as to inhibit the failure propagation of network nodes and improve the traffic network resilience. Therefore, the network resilience can be restored to the initial 100% state with sufficient time. In addition, with the improvement of the recovery rate, the time cost is greatly reduced.

The network resilience of the node recovery based on forecasting under different time advances is shown in Figure 17 and Table 8.

It can be seen from Figure 17 and Table 8 that with the decrease of the time forecasting advance, the network resilience bearing capacity and the overall network resilience continue to increase. The reduction of the time forecasting advance means that the security risks of the traffic network resilience can be discovered earlier, and the recovery strategy can be adopted earlier. As the earlier intervention of the recovery, the network node failure propagation can be greatly inhibited, the deterioration degree of the network resilience can be greatly slowed down, and the resilience bearing capacity can be significantly improved. However, the earlier intervention caused by the reduction of time forecasting advance has little effect on the final recovery degree of the network, and the resilience recovery capacity is almost unchanged. In addition, time costs continue to decrease as problems are identified earlier and optimization measures are taken.

Further, the results of network resilience optimization with and without node risk values are analyzed, as shown in Figure 18 and Table 9 below. It can be seen from the figure and table that the network resilience results have been effectively improved after combining the node risk value.

The computational complexity of this method is O(n). Te scalability of the network resilience optimization method under different network scales is analyzed, as shown in Table 10 below. It can be seen that when the network scale ranges from 20 nodes to 8000 nodes, the method can effectively improve the network resilience, and the increase in algorithm simulation experiment time is also within a reasonable range, which proves the effectiveness and scalability of the method for different network scales.

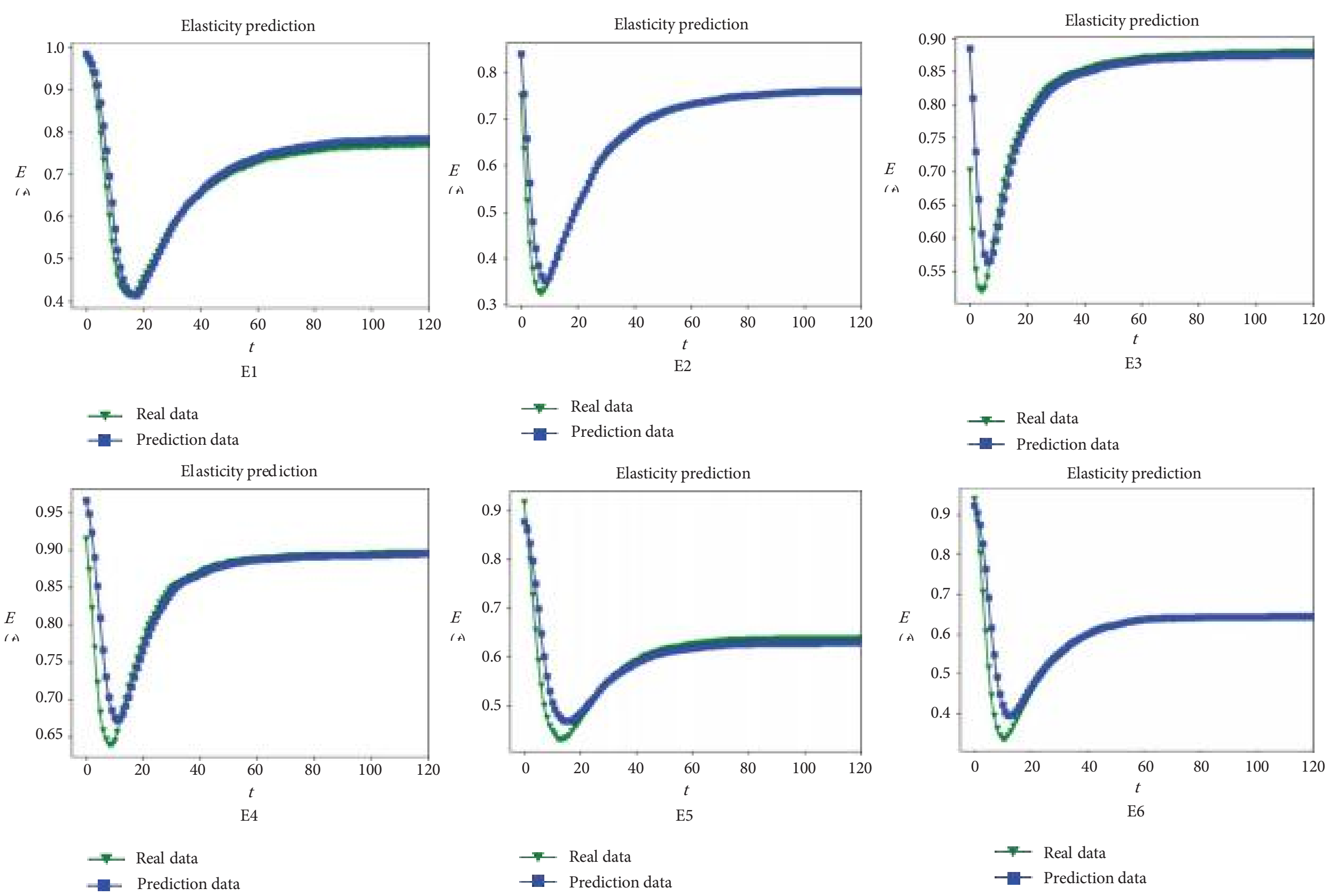

FIGURE 14: Forecasting results of traffic network resilience under different conditions.

TABLE 5: Forecasting accuracy of traffic network resilience under different conditions.

| | MRE | MSE | R2 coefficient |
|---|---|---|---|
| E1: 0.3, 0.1, 0.02 | 0.00639 | 0.00016 | 0.96842 |
| E2: 0.5, 0.1, 0.02 | 0.00593 | 0.00020 | 0.97014 |
| E3: 0.5, 0.3, 0.02 | 0.01057 | 0.00048 | 0.87830 |
| E4: 0.3, 0.3, 0.02 | 0.00666 | 0.00028 | 0.87052 |
| E5: 0.3, 0.1, 0.05 | 0.02395 | 0.00052 | 0.88427 |
| E6: 0.5, 0.1, 0.05 | 0.02560 | 0.00110 | 0.87663 |

—— Defense, defense = 0. 3 —— Defense, defense = 0. 5

FIGuRE 15: Trafc network resilience under diferent protection rates.

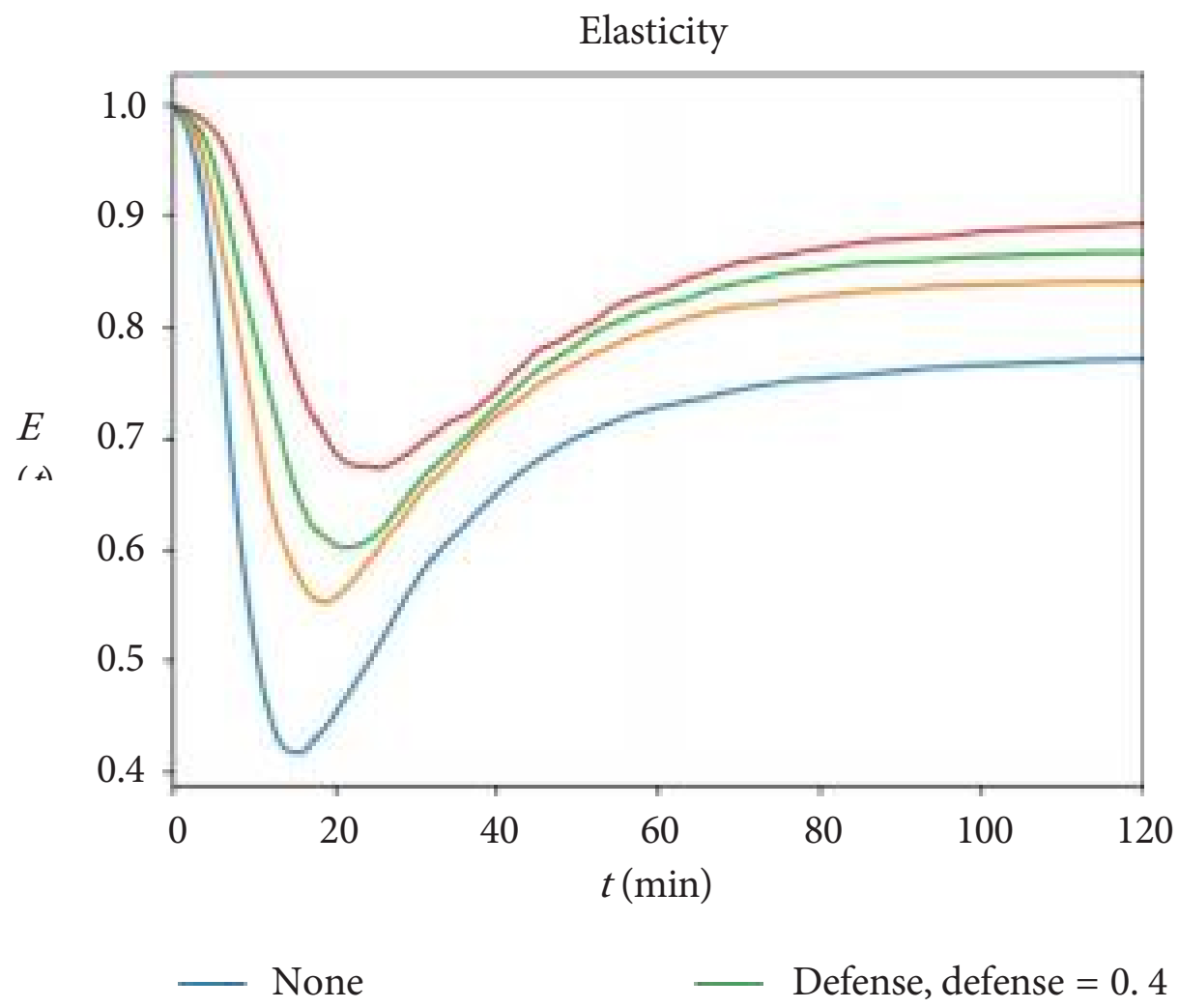

Defense, defense = 0. 3 Defense, defense = 0. 5

FIGuRE 15: Trafc network resilience under diferent protection rates.

TABLE 6: Traffic network resilience under different protection rates.

| Protection rates | $E_{withstand}$ | $E_{recover}$ | $E$ | Time cost (%) |
|---|---|---|---|---|
| No protection | 0.4241 | 0.7666 | 0.7310 | 100 |
| 0.3 | 0.5423 | 0.8470 | 0.8143 | 94.9 |
| 0.4 | 0.5928 | 0.8792 | 0.8572 | 79.5 |
| 0.5 | 0.6470 | 0.8952 | 0.8796 | 72.2 |

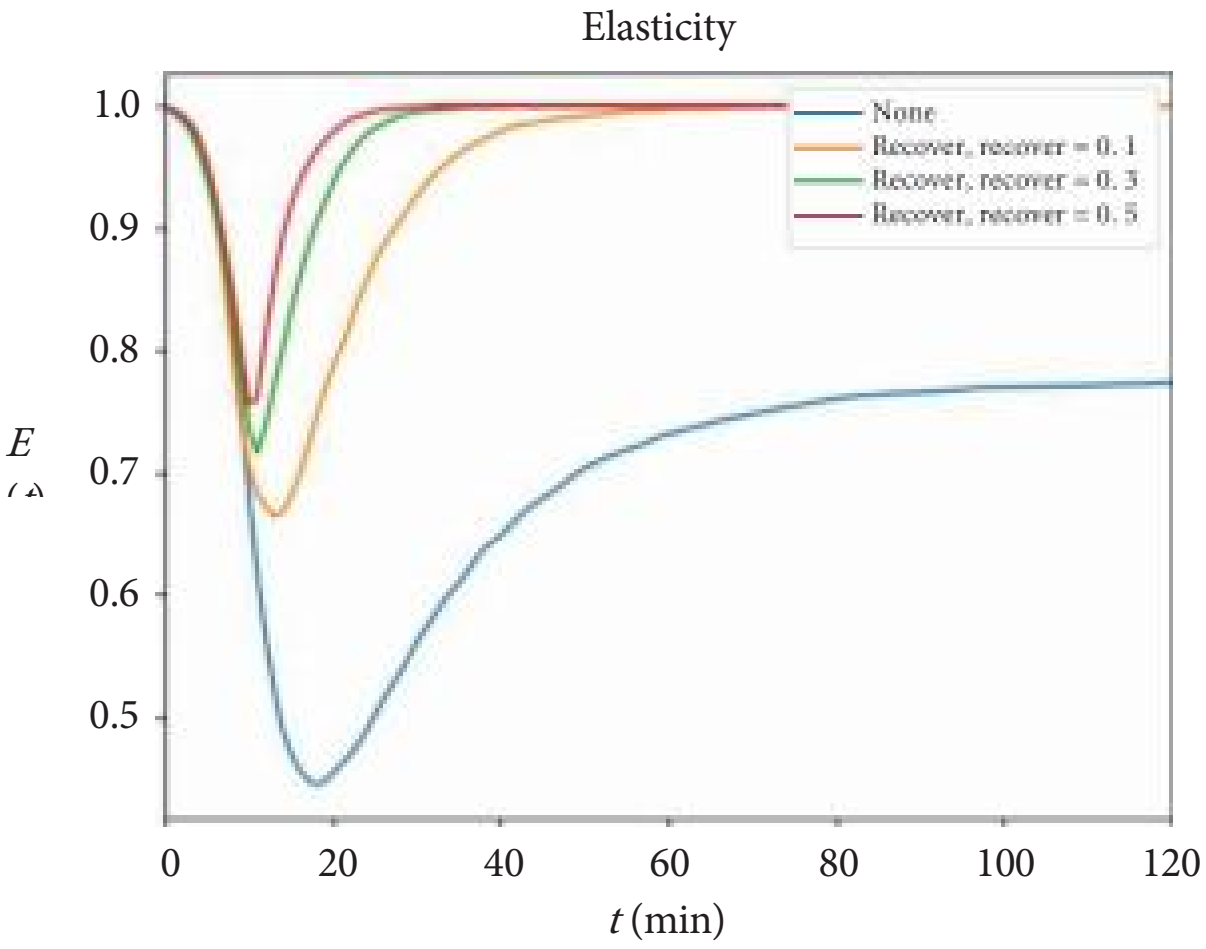

FIGURE 16: Traffic network resilience under different recovery rates.

TABLE 7: Traffic network resilience under different recovery rates.

| Recovery rates | $E_{withstand}$ | $E_{recover}$ | $E$ | Time cost (%) |
|---|---|---|---|---|
| No recovery | 0.3996 | 0.7668 | 0.7282 | 100 |
| 0.1 | 0.6310 | 0.9992 | 0.8942 | 41.1 |
| 0.3 | 0.7168 | 0.9998 | 0.8999 | 22.9 |
| 0.5 | 0.7494 | 1.0000 | 0.9117 | 16.4 |

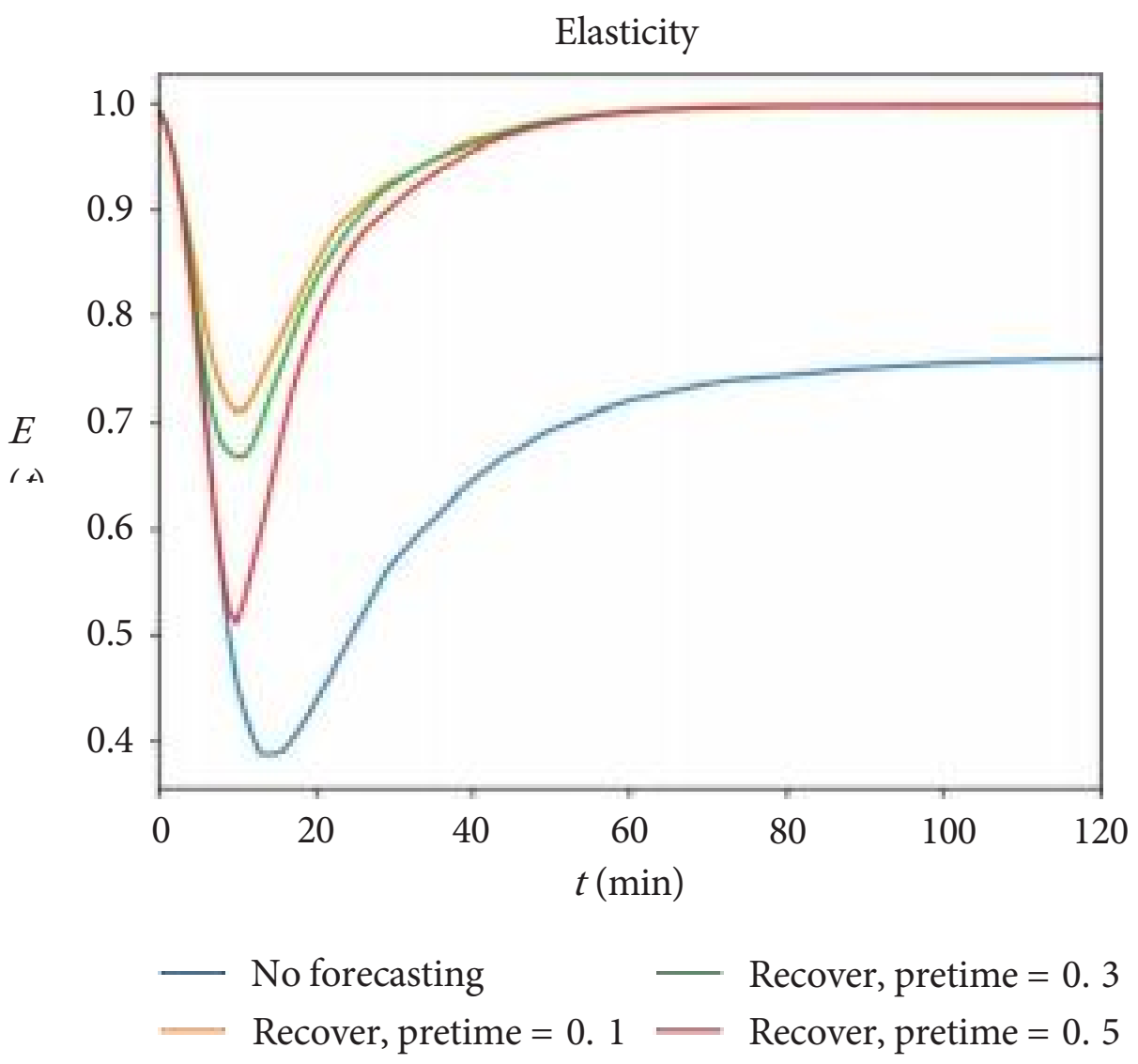

FIGURE 17: Traffic network resilience under different time advance.

TABLE 8: Traffic network resilience under different time advance.

| Time forecasting advance | $E_{withstand}$ | $E_{recover}$ | $E$ | Time cost (%) |
|---|---|---|---|---|
| No forecasting | 0.4094 | 0.7676 | 0.7261 | 100 |
| 0.5 | 0.5309 | 0.9982 | 0.9278 | 52.8 |
| 0.3 | 0.6748 | 0.9982 | 0.9354 | 38.6 |
| 0.1 | 0.7490 | 0.9982 | 0.9559 | 31.7 |

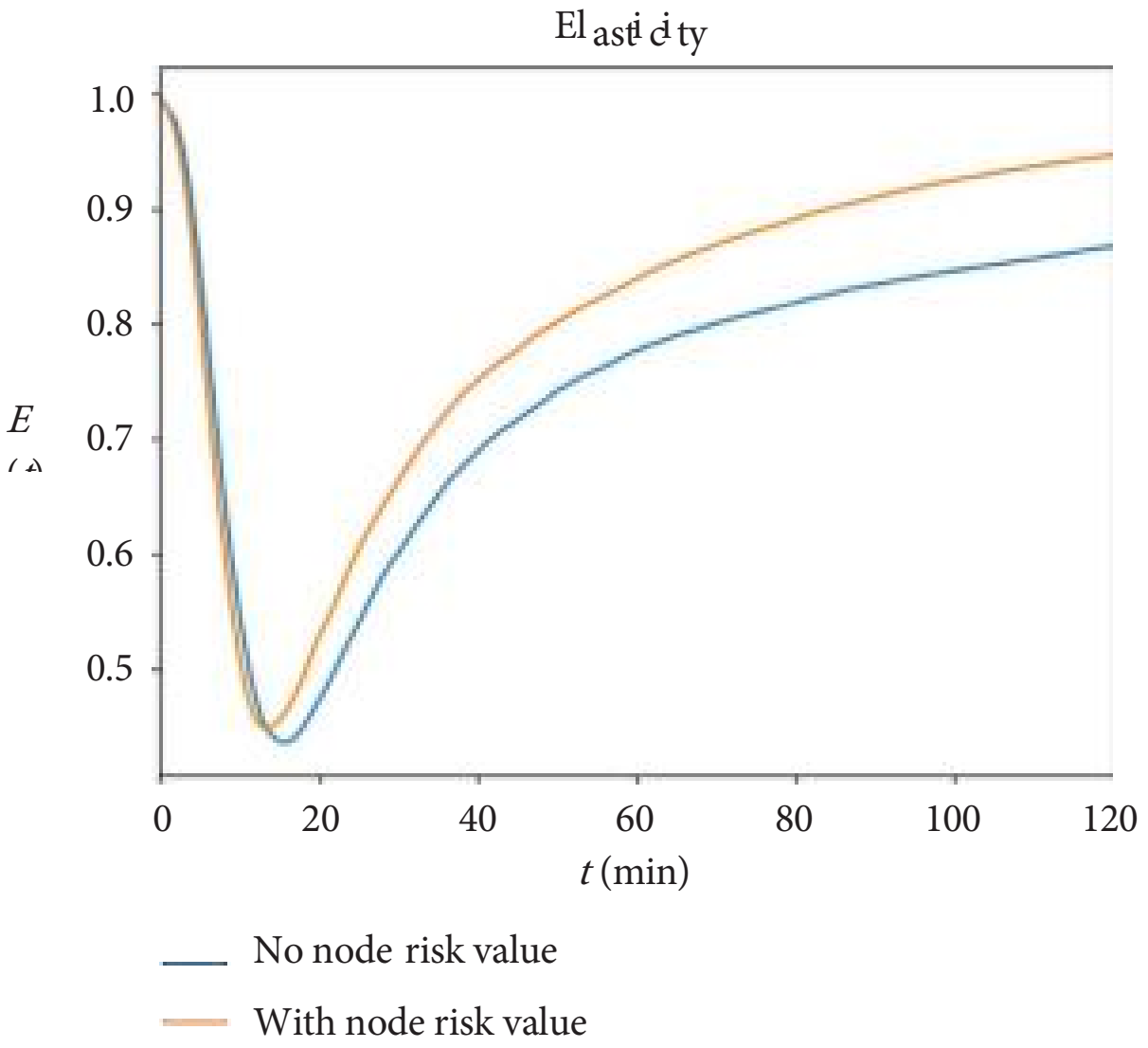

FIGURE 18: Traffic network resilience with or without node risk values.

TABLE 9: Traffic network resilience with or without node risk values.

| | $E_{withstand}$ | $E_{recover}$ | $E$ |
|---|---|---|---|
| With node risk value | 0.442 | 0.9902 | 0.8976 |
| No node risk value | 0.4116 | 0.9318 | 0.8103 |

TABLE 10: Resilience optimization of different network scales.

| Network scale (number of nodes) | Unoptimized | After-optimization | Simulation time |
|---|---|---|---|
| 20 | 0.6872 | 0.8972 | 0.23 |
| 50 | 0.6895 | 0.9010 | 0.24 |
| 100 | 0.6942 | 0.9017 | 0.30 |
| 300 | 0.7057 | 0.9216 | 0.35 |
| 800 | 0.7033 | 0.9124 | 0.37 |
| 1000 | 0.7261 | 0.9354 | 0.38 |
| 3000 | 0.7141 | 0.9121 | 0.80 |
| 5000 | 0.7434 | 0.9538 | 1.30 |
| 8000 | 0.7762 | 0.9615 | 2.37 |

Ten, the network resilience optimization experiments are carried out in aviation networks, industrial networks, power networks, and other complex networks. The network The resilience optimization method is based on trend forecasting and combines the node protection and node recovery strategies. The results are shown in Table 11.

TABLE 11: Resilience optimization of different networks.

| | Unoptimized | After-optimization |
|---|---|---|
| Traffic network | 0.7261 | 0.9354 |
| Aviation network | 0.7389 | 0.9519 |
| Industrial network | 0.6748 | 0.9186 |
| Power network | 0.7440 | 0.9263 |

As can be seen from Table 11, the network resilience optimization method proposed in this paper can achieve good results in a variety of complex network applications, which proves the effectiveness, scalability, and adaptability of the method.

## 5. Conclusion

With the ongoing advancement of computer technology, digitalization and networking have emerged as prevailing trends in contemporary society. However, this development has led to a progressively intricate traffic infrastructure that poses numerous security challenges. Aiming at the incompleteness and inadequacy of the resilience modeling, forecasting, and recovery of complex traffic networks, the following research is carried out in this paper. This paper conducts an initial analysis of the failure propagation process of network nodes following a traffic network attack, and introduces the risk value to establish the SIRD-R failure propagation model. Then, the resilience bearing capacity and resilience recovery capacity of the traffic network are analyzed comprehensively, and the network resilience model of the traffic network is established from two aspects: real-time resilience and overall resilience. Furthermore, the traffic network's resilience is forecasted using LSTM networks to enable advanced perception and early warning capabilities. Finally, based on the resilience forecasting of the traffic network, the resilience recovery strategy of the traffic network based on the forecasting is proposed to realize the proactive optimization of the traffic network resilience in advance. This study aims to disrupt the causal chain of fault diffusion within the network, enhance the resilience of complex traffic networks, and bolster the protective capacity of traffic networks against potential attacks. Ultimately, this will enable effective network security emergency response, ensure uninterrupted network operation, and mitigate economic losses or major accidents.

## Data Availability Statement

The datasets used in this paper are Gold Coast, a city in southern Queensland, Australia, which can be downloaded at https://github.com/bstabler/TransportationNetworks.

## Conflicts of Interest

The authors declare no conflicts of interest.

## Funding

Tis work was supported by Beijing Natural Science Foundation (L222005); National Key Research and Development Program (2022YFB3103602); CCF-NSFOCUS Kun-Peng Scientific Research Fund (CCF-NSFO-CUS202215); Industrial Internet Innovation and Development Project (TC210804R) and (CEIEC-2022-ZM02-0252); and Fundamental Research Funds for the Central Universities.